\documentclass[fleqn,10pt]{wlscirep}
\usepackage[utf8]{inputenc}
\usepackage[T1]{fontenc}

\usepackage{subfig}
\title{Dynamic Reconfiguration of Robotic Swarms:  Coordination and Control for Precise Shape Formation}

\author{Prab Prasertying$^{1}$, Paulo Garcia$^{2}$ and Warisa Sritriratanarak$^1$}
\affil{$^1$Department of Computer Engineering, Chulalongkorn University, Bangkok, Thailand}
\affil{$^2$International School of Engineering, Chulalongkorn University, Bangkok, Thailand}
\affil{prab.p@chula.ac.th, paulo.g@chula.ac.th, warisa.s@chula.ac.th}

\begin{abstract}
Coordination of movement and configuration in robotic swarms is a challenging endeavor.
Deciding when and where each individual robot must move is a computationally complex problem. The challenge is further exacerbated by difficulties inherent to physical systems, such as measurement error and control dynamics.
Thus, how to best determine the optimal path for each robot, when moving from one configuration to another, and how to best perform such determination and effect corresponding motion remains an open problem.
In this paper, we show an algorithm for such coordination of robotic swarms.
Our methods allow seamless transition from one configuration to another, leveraging geometric formulations that are mapped to the physical domain through appropriate control, localization, and mapping techniques.
This paves the way for novel applications of robotic swarms by enabling more sophisticated distributed behaviors.
\end{abstract}
\begin{document}

\flushbottom
\maketitle
\thispagestyle{empty}

\section*{Introduction}

\par Mobile robotic swarms \cite{turgut2008self} are increasingly being deployed in a variety of critical and dynamic environments \cite{peleg2005distributed}. Their ability to operate autonomously, adapt to unknown terrain, and function collectively makes them ideal for tasks such as environmental monitoring \cite{blender2016managing}, search and rescue, planetary exploration, and infrastructure inspection \cite{schranz2020swarm}. In these scenarios, large numbers of simple robots are often preferred over a single complex machine due to their scalability, fault tolerance, and parallelism \cite{arvin2018perpetual}.

\par However, a major challenge in leveraging swarm robotics lies in coordinating individual robots to act as a coherent group \cite{peleg2005distributed}. For example, in search and rescue operations, robots may need to form communication relays or sweep a region efficiently without duplication. In structural assembly, such as deploying solar panels or temporary shelters, robots must move in a carefully orchestrated sequence to avoid collisions and gaps. This coordination becomes especially complex when targeting precise formation shapes or when operating under limited communication and sensing capabilities.

\par To address these challenges, a range of coordination methods have been proposed. SLAM (Simultaneous Localization and Mapping \cite{kegeleirs2021swarm}) enables individual robots to build and navigate maps of unknown environments, while GroupSLAM extends this to multiple robots working cooperatively. In addition, distributed path planning algorithms such as consensus-based control, potential fields, and virtual forces have enabled robust multi-agent motion. Nonetheless, most of these techniques are optimized for navigation and coveragel; not for precise, collective shape formation.

\par This paper builds upon the Centralized Planning and Distributed Execution framework introduced by Liu et al. \cite{liu2025centralizedplanningdistributedexecution}, which allows a global planner to assign collision-free motion paths -called ribbons- to each robot on a lattice. Robots then execute these paths using only local neighbor information, enabling accurate shape formation with minimal communication overhead. While effective, this approach leaves open questions about how well the system performs under real-world uncertainty, such as sensor noise, robot failure, and communication loss.

\section{Background}

\subsection{Probabilistic SLAM Formulation}

Let $\mathbf{x}_{0:t} = \{\mathbf{x}_0, \mathbf{x}_1, \ldots, \mathbf{x}_t\}$ denote the robot's trajectory and $\mathbf{m} = \{\mathbf{m}_1, \ldots, \mathbf{m}_M\}$ represent $M$ landmarks in the environment. Given control inputs $\mathbf{u}_{1:t}$ and sensor observations $\mathbf{z}_{1:t}$, SLAM estimates the joint posterior:

\begin{equation}
p(\mathbf{x}_{0:t}, \mathbf{m} \mid \mathbf{z}_{1:t}, \mathbf{u}_{1:t})
\end{equation}

This can be factored using Bayes' rule and the Markov assumption:

\begin{equation}
\begin{aligned}
p(\mathbf{x}_{0:t}, \mathbf{m} \mid \mathbf{z}_{1:t}, \mathbf{u}_{1:t}) \propto \, &p(\mathbf{x}_0) \prod_{k=1}^{t} p(\mathbf{x}_k \mid \mathbf{x}_{k-1}, \mathbf{u}_k) \\
&\times \prod_{k=1}^{t} p(\mathbf{z}_k \mid \mathbf{x}_k, \mathbf{m})
\end{aligned}
\end{equation}

where:
\begin{itemize}
    \item $p(\mathbf{x}_k \mid \mathbf{x}_{k-1}, \mathbf{u}_k)$ is the motion model
    \item $p(\mathbf{z}_k \mid \mathbf{x}_k, \mathbf{m})$ is the observation model
\end{itemize}

\subsection{Extended Kalman Filter SLAM}

EKF-SLAM maintains a Gaussian belief over the state vector $\mathbf{X} = [\mathbf{x}^T, \mathbf{m}^T]^T$:

\begin{equation}
\mathbf{X} \sim \mathcal{N}(\boldsymbol{\mu}, \boldsymbol{\Sigma})
\end{equation}

The EKF prediction step linearizes the motion model:

\begin{equation}
\begin{aligned}
\boldsymbol{\mu}_{t|t-1} &= g(\boldsymbol{\mu}_{t-1}, \mathbf{u}_t) \\
\boldsymbol{\Sigma}_{t|t-1} &= G_t \boldsymbol{\Sigma}_{t-1} G_t^T + R_t
\end{aligned}
\end{equation}

where $G_t = \frac{\partial g}{\partial \mathbf{X}}\big|_{\boldsymbol{\mu}_{t-1}}$ is the Jacobian and $R_t$ is the process noise covariance.

The update step incorporates observations:

\begin{equation}
\begin{aligned}
\mathbf{K}_t &= \boldsymbol{\Sigma}_{t|t-1} H_t^T (H_t \boldsymbol{\Sigma}_{t|t-1} H_t^T + Q_t)^{-1} \\
\boldsymbol{\mu}_t &= \boldsymbol{\mu}_{t|t-1} + \mathbf{K}_t (\mathbf{z}_t - h(\boldsymbol{\mu}_{t|t-1})) \\
\boldsymbol{\Sigma}_t &= (I - \mathbf{K}_t H_t) \boldsymbol{\Sigma}_{t|t-1}
\end{aligned}
\end{equation}

where $H_t = \frac{\partial h}{\partial \mathbf{X}}\big|_{\boldsymbol{\mu}_{t|t-1}}$ and $Q_t$ is the measurement noise covariance.

\section{Robot Motion Model}

We model each robot as a differential-drive mobile platform operating in a 2D workspace $\mathcal{W} \subset \mathbb{R}^2$. The configuration of robot $i$ at time $t$ is described by its pose:

\begin{equation}
\mathbf{q}_i(t) = [x_i(t), y_i(t), \theta_i(t)]^T
\end{equation}

where $(x_i, y_i)$ represents the robot's position in the global frame and $\theta_i \in [-\pi, \pi)$ is its orientation.

The kinematic model for a differential-drive robot is given by:

\begin{equation}
\begin{bmatrix}
\dot{x}_i \\
\dot{y}_i \\
\dot{\theta}_i
\end{bmatrix}
=
\begin{bmatrix}
\cos\theta_i & 0 \\
\sin\theta_i & 0 \\
0 & 1
\end{bmatrix}
\begin{bmatrix}
v_i \\
\omega_i
\end{bmatrix}
\end{equation}

where $v_i$ is the linear velocity and $\omega_i$ is the angular velocity. These control inputs are constrained by:

\begin{equation}
|v_i| \leq v_{\max}, \quad |\omega_i| \leq \omega_{\max}
\end{equation}

Accurate localization is critical for multi-robot coordination. We employ a sensor fusion approach combining GPS and odometry for localization and state Estimation.

Wheel encoders provide incremental pose updates. Given encoder readings at time $t$ and $t-1$, the odometry-based pose update is:

\begin{equation}
\begin{aligned}
\Delta s &= \frac{r}{2}(\Delta\phi_R + \Delta\phi_L) \\
\Delta\theta &= \frac{r}{L}(\Delta\phi_R - \Delta\phi_L) \\
x_i(t) &= x_i(t-1) + \Delta s \cos(\theta_i(t-1) + \Delta\theta/2) \\
y_i(t) &= y_i(t-1) + \Delta s \sin(\theta_i(t-1) + \Delta\theta/2) \\
\theta_i(t) &= \theta_i(t-1) + \Delta\theta
\end{aligned}
\end{equation}

where $r$ is the wheel radius, $L$ is the wheelbase, and $\Delta\phi_R$, $\Delta\phi_L$ are the right and left wheel angular displacements.

Odometry suffers from cumulative drift due to wheel slippage and uneven terrain. The position error grows unbounded:

\begin{equation}
\sigma_{x,y}^2(t) \approx k_1 d(t) + k_2 |\Delta\theta|(t)
\end{equation}

where $d(t)$ is the distance traveled and $k_1, k_2$ are error coefficients.

GPS provides absolute position measurements with bounded error:

\begin{equation}
\mathbf{z}_{\text{GPS}} = \mathbf{p}_i + \mathbf{n}_{\text{GPS}}
\end{equation}

where $\mathbf{p}_i = [x_i, y_i]^T$ is the true position and $\mathbf{n}_{\text{GPS}} \sim \mathcal{N}(0, \sigma_{\text{GPS}}^2 \mathbf{I})$ with associated $\sigma_{\text{GPS}}$ .

We fuse odometry and GPS using a complementary filter. The fused position estimate is:

\begin{equation}
\begin{aligned}
\hat{x}_i(t) &= \alpha \cdot x_{\text{GPS}}(t) + (1-\alpha) \cdot x_{\text{odom}}(t) \\
\hat{y}_i(t) &= \alpha \cdot y_{\text{GPS}}(t) + (1-\alpha) \cdot y_{\text{odom}}(t)
\end{aligned}
\end{equation}

where $\alpha \in [0,1]$ is a tuning parameter.

\subsection{Communication}
The communication topology among $N$ robots is modeled as a time-varying graph $\mathcal{G}(t) = (\mathcal{V}, \mathcal{E}(t))$, where $\mathcal{V} = \{1, \ldots, N\}$ and $\mathcal{E}(t) \subseteq \mathcal{V} \times \mathcal{V}$.

An edge $(i,j) \in \mathcal{E}(t)$ exists if:

\begin{equation}
\|\mathbf{p}_i(t) - \mathbf{p}_j(t)\| \leq r_{\text{comm}}
\end{equation}

where $r_{\text{comm}}$ is the communication range. In our system, we assume global communication ($r_{\text{comm}} = \infty$), though local channels with specific $r_{\text{comm}}$ are also available.

\subsubsection{Collision Avoidance via Velocity Obstacles}

For safe navigation, robot $i$ must avoid collisions with robot $j$. The velocity obstacle $VO_{i|j}$ is defined as:

\begin{equation}
VO_{i|j} = \{\mathbf{v}_i \mid \exists t > 0 : \mathbf{p}_i + t\mathbf{v}_i \in B(\mathbf{p}_j + t\mathbf{v}_j, r_i + r_j)\}
\end{equation}

where $B(\mathbf{c}, r)$ is a disk of radius $r$ centered at $\mathbf{c}$, and $r_i, r_j$ are robot radii.

Robot $i$ selects a collision-free velocity:

\begin{equation}
\mathbf{v}_i^* = \arg\min_{\mathbf{v}_i \notin \bigcup_j VO_{i|j}} \|\mathbf{v}_i - \mathbf{v}_i^{\text{pref}}\|
\end{equation}

where $\mathbf{v}_i^{\text{pref}}$ is the preferred velocity toward the goal.

\subsubsection{Formation Control}

For shape formation, robots must converge to a desired configuration $\mathbf{q}_i^d$. A common control law is:

\begin{equation}
\mathbf{u}_i = -K_p (\mathbf{q}_i - \mathbf{q}_i^d) - K_d \dot{\mathbf{q}}_i
\end{equation}

where $K_p, K_d$ are proportional and derivative gains. Stability is guaranteed if the gains satisfy:

\begin{equation}
K_p > 0, \quad K_d > 2\sqrt{K_p}
\end{equation}

\subsection{Distributed Consensus}

In distributed systems, robots reach agreement on shared variables through consensus protocols. Let $x_i(t)$ be robot $i$'s estimate of a consensus variable. The discrete-time consensus update is:

\begin{equation}
x_i(t+1) = x_i(t) + \epsilon \sum_{j \in \mathcal{N}_i(t)} (x_j(t) - x_i(t))
\end{equation}

where $\mathcal{N}_i(t)$ is the set of neighbors and $\epsilon > 0$ is the step size. Convergence to global consensus $\lim_{t \to \infty} x_i(t) = \bar{x}$ is guaranteed if $\mathcal{G}(t)$ is connected and $\epsilon < 1/d_{\max}$, where $d_{\max}$ is the maximum degree.

\section{Row-Based Formation with Distributed Completion Tracking}

We consider a swarm of $N$ homogeneous mobile robots tasked with filling a target shape $\mathcal{S}$ defined on a 2D grid with cell size $c$. The shape is represented as a set of target positions:

\begin{equation}
\mathcal{S} = \{(x_i, y_i) \mid i = 1, 2, \ldots, |\mathcal{S}|\}
\end{equation}

Each robot $r_j$ (where $j \in \{0, 1, \ldots, N-1\}$) has:
\begin{itemize}
    \item GPS positioning
    \item Omnidirectional radio communication (global channel with unlimited range, local channel with  specific range)
    \item Differential-drive locomotion with maximum velocity $v_{max}$
    \item Odometry with encoder feedback for dead reckoning
\end{itemize}

The goal is to assign each robot to a unique position in $\mathcal{S}$ and navigate collision-free while minimizing completion time.

\subsection{Side Assignment and Row Partitioning}

Unlike traditional approaches that use a single gradient from a seed robot, we partition the target shape into \textbf{rows} of $k = 6$ robots each. Rows are assigned to either the LEFT or RIGHT side based on the row number's parity:

\begin{equation}
\text{side}(n) = \begin{cases}
\text{RIGHT} & \text{if } n \bmod 2 = 1 \\
\text{LEFT} & \text{if } n \bmod 2 = 0
\end{cases}
\end{equation}

For each side, target positions are sorted by distance from the shape's center $\mathbf{c} = (\bar{x}, \bar{y})$:

\begin{equation}
d_i = \sqrt{(x_i - \bar{x})^2 + (y_i - \bar{y})^2}
\end{equation}

Robots in each row are assigned targets in \textbf{center-out order}, ensuring inner positions fill first. This creates two independent filling streams that work in parallel. Each robot executes a three-phase motion plan:

\subsubsection{Phase 0: Early Position Check}

Before any movement, robot $r_j$ checks if its current position $\mathbf{p}_j$ matches any target position in $\mathcal{S}$:

\begin{equation}
\exists\, (x_i, y_i) \in \mathcal{S} : \|\mathbf{p}_j - (x_i, y_i)\| \leq \epsilon
\end{equation}

 If matched:
\begin{enumerate}
    \item Broadcast \texttt{OCCUPIED} message with coordinates
    \item Broadcast \texttt{ROW\_ROBOT\_DONE} message with row number
    \item If $r_j$ is the last robot in its row and all 6 robots are done, broadcast \texttt{ROW\_MOVING}
    \item Skip to formation complete
\end{enumerate}

This optimization handles cases where robots are pre-positioned at their targets.

\subsubsection{Phase 1: Starting Point Navigation}

Each side has a \textbf{starting point line} offset from the target area edge:

\begin{equation}
\begin{aligned}
x_{\text{start}}^{\text{RIGHT}} &= \max_{(x,y) \in \mathcal{S}} x + 0.25 \\
x_{\text{start}}^{\text{LEFT}} &= \min_{(x,y) \in \mathcal{S}} x - 0.25
\end{aligned}
\end{equation}

The starting point for robot $r_j$ assigned to target $(x_t, y_t)$ is:

\begin{equation}
\mathbf{s}_j = (x_{\text{start}}^{\text{side}(n)}, y_t)
\end{equation}

\textbf{Row Coordination Rule}: Row $n$ can only begin Phase 1 if row $n-2$ has completed Phase 2 (final target arrival). This prevents traffic congestion by ensuring odd rows move sequentially (1 $\rightarrow$ 3 $\rightarrow$ 5) and even rows move sequentially (2 $\rightarrow$ 4 $\rightarrow$ 6).

Navigation uses an \textbf{axis-aligned controller}: Y-axis first, then X-axis, with odometry + GPS fusion for positioning.

\subsubsection{Phase 2: Final Target Navigation with Staggered Delays}

Upon reaching the starting point, robot $r_j$ with arrival order $o_j$ waits:

\begin{equation}
t_{\text{delay}} = 1.0 + (o_j \times 3.0) \text{ seconds}
\end{equation}

where $o_j \in \{0, 1, 2, 3, 4, 5\}$ based on arrival sequence at the starting line. This staggered release prevents collisions during final approach.

Before moving, $r_j$ performs a second position check (Phase 0 late check) similar to the early check.

Once the delay expires, $r_j$ navigates to its target $(x_t, y_t)$ using the same axis-aligned controller.

\subsection{Distributed Completion Tracking}

To coordinate across rows, we use a distributed completion tracking mechanism:

\textbf{State Variables} (maintained by each robot):
\begin{itemize}
    \item \texttt{completed\_robots[n]}: Set of robot names that finished in row $n$
    \item \texttt{can\_my\_row\_move}: Boolean flag indicating permission to start Phase 1
    \item \texttt{row\_start\_broadcast}: Boolean flag to prevent duplicate \texttt{ROW\_MOVING} messages
\end{itemize}

\textbf{Message Protocol}:

\begin{enumerate}
    \item \textbf{ROW\_ROBOT\_DONE}: When robot $r_j$ in row $n$ reaches its target:\\
    $\text{broadcast}(\texttt{ROW\_ROBOT\_DONE}, n, \text{name}_j)$\\
    All robots update: $\texttt{completed\_robots}[n] \leftarrow \texttt{completed\_robots}[n] \cup \{\text{name}_j\}$

    \item \textbf{ROW\_MOVING}: The last robot in row $n$ (robot with order 5) periodically checks:\\
    $|\texttt{completed\_robots}[n]| \geq k$\\
    If true, broadcast: $\text{broadcast}(\texttt{ROW\_MOVING}, n)$\\
    This unlocks row $n+2$ by setting $\texttt{can\_my\_row\_move} \leftarrow \text{True}$ for all robots in that row.
\end{enumerate}

\textbf{Periodic Verification}: To handle message loss or out-of-order arrivals, the last robot checks completion every 1 second after reaching its target, ensuring \texttt{ROW\_MOVING} is eventually sent even if some robots finish late. Collisions are avoided through multiple mechanisms: 

\begin{enumerate}
    \item \textbf{Spatial Separation}: LEFT and RIGHT sides operate on opposite edges of the target area
    \item \textbf{Row Serialization}: Rows on the same side wait for previous row completion
    \item \textbf{Staggered Release}: 3-second delays between robots in the same row
    \item \textbf{Target Reservation}: \texttt{CLAIM} messages prevent multiple robots from selecting the same target
    \item \textbf{OCCUPIED Broadcast}: Robots announce their final positions to prevent conflicts
\end{enumerate}

\section{Experiments and Results}

\begin{figure}[htp]
\subfloat[]{\includegraphics[width=.3\textwidth]{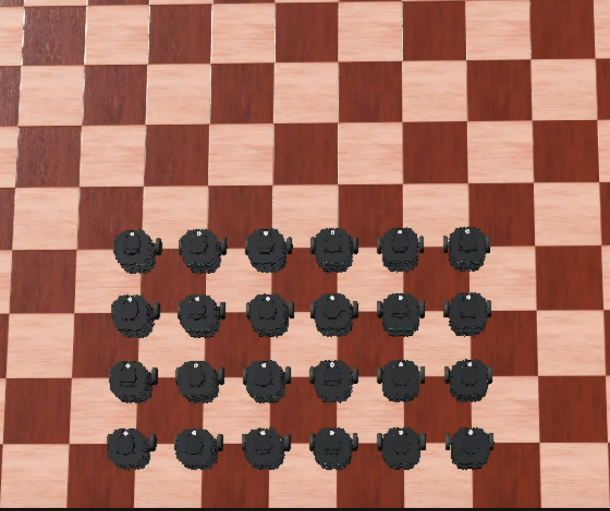}}\quad
\subfloat[]{\includegraphics[width=.3\textwidth]{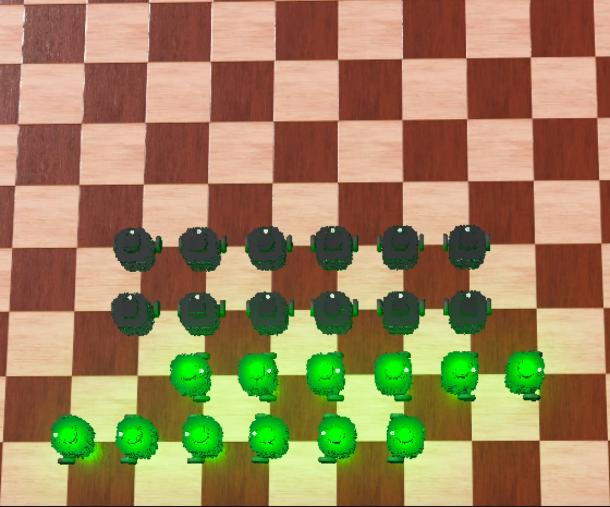}}\quad
\subfloat[]{\includegraphics[width=.3\textwidth]{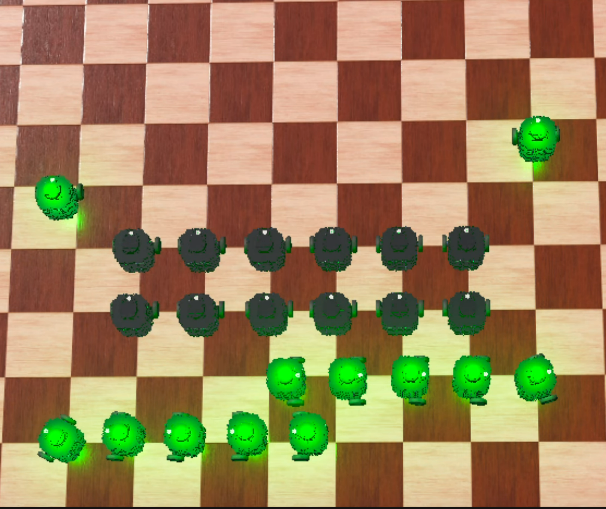}}\\
\subfloat[]{\includegraphics[width=.3\textwidth]{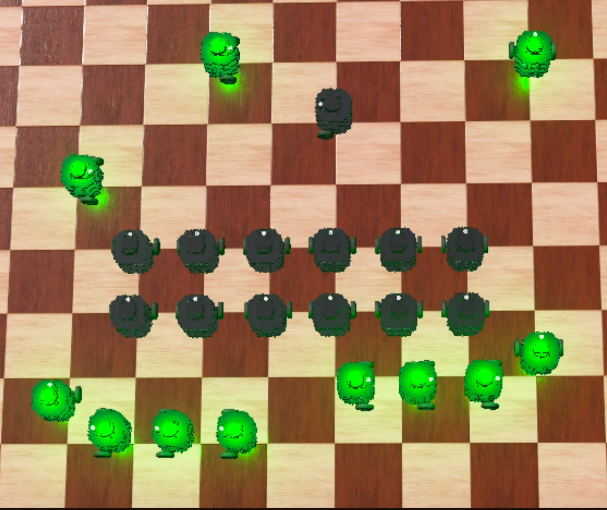}}\quad
\subfloat[]{\includegraphics[width=.3\textwidth]{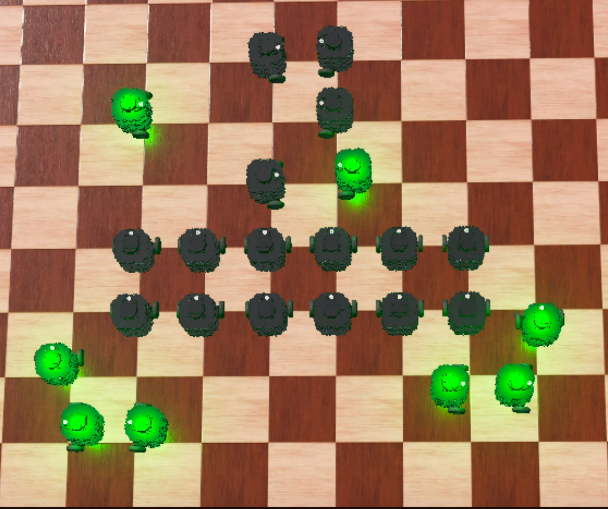}}\quad
\subfloat[]{\includegraphics[width=.3\textwidth]{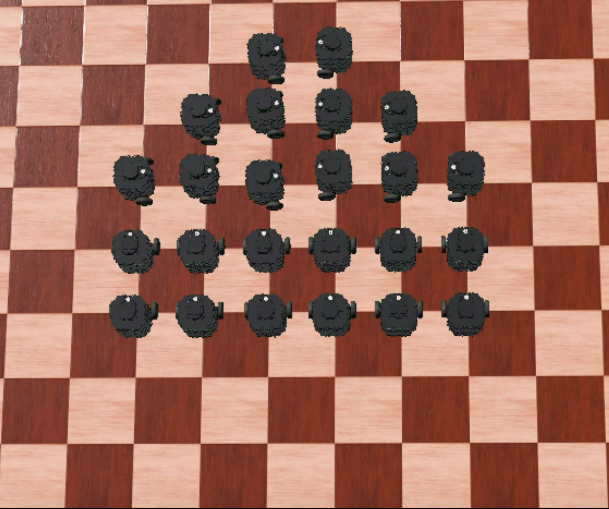}}
\label{fig:results}
\caption{Simulated robotic swarm movement when changing from rectangular configuration (a) to arrowhead configuration (f). Intermediate steps sequentially from (b) to (e). Full video of simulation available at \url{https://youtu.be/TrFtKlr0_5M}.}
\end{figure}

We simulate robotic motion control on Webots with TurtleBot3 Burger robots, using the following parameters: $v_{\max} = 0.22$ m/s, $\omega_{\max} = 2.84$ rad/s, $r = 0.033$ m, $L = 0.16$ m, $c = 0.25$ meters, and $r_{\text{comm}} = 0.2$ m. In our implementation, $\alpha = 0.7$ weights GPS more heavily to prevent drift accumulation, given GPS positioning with accuracy $\pm 5$ cm ($\sigma_{\text{GPS}} = 0.05$ m.), Omnidirectional radio communication (global channel with unlimited range, local channel with 0.2m range), Differential-drive locomotion with maximum velocity $v_{max} = 0.22$ m/s, and $\epsilon = 0.05$ m (5 cm tolerance). Simulation results are depicted in Fig. 1.

\section{Related Work}

\par Swarm robotic systems aim to coordinate many simple robots to form complex structures or perform collective tasks. While traditional coordination techniques such as SLAM and distributed path planning focus on navigation and mapping, shape formation presents unique challenges: robots must not only avoid collisions, but also fill specific geometries with high accuracy, resilience, and scalability. Below, we summarize three foundational approaches that have shaped the field and define the trajectory of this paper. Rubenstein et al. \cite{Rubenstein2014795} demonstrated a fully decentralized shape formation algorithm using over 1,000 Kilobots. Their approach relied on simple local interactions: robots established a gradient field from a seed, performed edge-following, and assigned roles based on local context. This method was highly scalable and robust to individual robot failure. However, it had limited shape precision, particularly near concavities and boundaries, and could not guarantee complete filling of complex structures. Additionally, convergence was slow due to the purely local nature of decision-making and absence of global oversight.
\par To improve both efficiency and precision, Yang et al. \cite{yang2022parallel} introduced a parallel and distributed self-assembly algorithm based on motion chains and stratified growth. Their system operated on a lattice grid, where robots filled a target shape layer by layer. Two synchronized "motion-chains" advanced in opposite directions, enabling parallel filling of the shape's outer layers. The approach provided collision avoidance and relatively fast convergence, while still relying solely on local communication and sensing. However, it assumed a static and clean environment and could be sensitive to robot dropout or environmental noise. The method also struggled with holes or narrow channels inside the target shape.
\par Liu et al. \cite{liu2025centralizedplanningdistributedexecution} presented a hybrid strategy that combines the strengths of centralized and decentralized systems. A centralized planner first partitions the target shape (defined on a hexagonal lattice) into ribbons: non-overlapping paths that each robot can follow to its destination. These motion plans are assigned offline and require minimal storage. Once deployed, robots execute their ribbon-based plans using only local neighbor distance sensing, making the system robust during execution even if communication fails. This method achieves high fill accuracy, fast convergence, and supports hole-rich or complex geometries. However, it assumes that a global planner can access and process the entire shape before execution begins, which may not be feasible in dynamic or real-time settings.
\par One of the earliest implementations of gradient-based control in modular robot assembly was proposed by Stoy \cite{Stoy}. In his work on self-reconfigurable robots, he introduced a method in which a seed robot emits a gradient value, and surrounding modules compute their own values based on the minimum of their neighbors plus one. This created a virtual scalar field that guided the growth of the structure through local interactions only.

\par This system used two distinct types of gradients: (1) A wander gradient, propagated by robots outside the desired shape. (2) A hole gradient, propagated by unfilled cells inside the shape. Robots used both to navigate: external robots would descend the hole gradient (toward unfilled areas) and ascend the wander gradient (away from congested zones), resulting in boundary movement and internal reshaping of the formation. This dual-gradient mechanism allowed the swarm to reconfigure itself into non-hollow, dense shapes from arbitrary initial conditions.

\section{Conclusions}

This paper presented a row-based coordination system for robotic swarm shape formation that combines centralized target assignment with distributed execution. Our approach enables efficient bilateral filling through three key mechanisms: row partitioning with side assignment, three-phase navigation with completion tracking, and multi-layered collision avoidance. 
\par Simulation experiments in Webots with TurtleBot3 Burger robots successfully demonstrated reconfiguration from rectangular to arrowhead formations. All robots completed the reconfiguration without collisions, validating both the coordination protocol and collision avoidance mechanisms. Full video documentation is available at \url{https://youtu.be/TrFtKlr0_5M}.
\par Our row-based approach occupies a practical position in the swarm formation design space. Compared to fully decentralized gradient-based methods \cite{Rubenstein2014795, Stoy}, we trade some adaptability for significantly improved predictability, easier debugging, and faster convergence through direct point-to-point navigation. Relative to centralized ribbon planning \cite{liu2025centralizedplanningdistributedexecution}, we sacrifice support for complex hole-filled geometries in exchange for implementation simplicity and straightforward verification. These trade-offs make our system particularly well-suited for applications requiring: (1) rapid prototyping and deployment in time-constrained scenarios, (2) predictable robot behavior for safety-critical applications, (3) simple to moderately complex convex or concave target shapes without internal holes, and (4) environments where GPS localization is available. The practical simplicity enables real-world deployment while maintaining formation precision within acceptable tolerance thresholds.

\bibliography{references}

\end{document}